\documentclass[letterpaper, 10 pt, conference]{ieeeconf}

\IEEEoverridecommandlockouts       

\usepackage{times} 
\usepackage{amsmath} 
\usepackage{amssymb}  
\usepackage{algorithm}
\usepackage[noend]{algpseudocode}
\usepackage{comment}
\usepackage{graphicx}
\usepackage{subfigure}
\usepackage{tikz}
\usepackage{etoolbox}
\usepackage{xspace}
\usepackage[letterspace=0,tracking=true]{microtype}
\usepackage{comment}
\usepackage{hyperref}
\usepackage[utf8]{inputenc}
\usepackage[T2A,T1]{fontenc}
\usepackage{booktabs}
\usepackage{paralist}

\makeatletter
\let\NAT@parse\undefined
\makeatother
\usepackage[numbers]{natbib}

\hypersetup{%
  pdfborder = {0 0 0},
  bookmarksdepth = section,
  citecolor = DarkGreen,
  linkcolor = DarkBlue,
  urlcolor = DarkBlue,
}%

\graphicspath{{images/}}

\definecolor{Red}{rgb}{1,0,0}
\definecolor{Green}{rgb}{0,0.69,0}
\definecolor{Blue}{rgb}{0,0,1}
\definecolor{LightBlue}{rgb}{0,0.5,1}
\definecolor{veryLightBlue}{rgb}{0.85,0.98,1}
\definecolor{veryLightGreen}{rgb}{0.6,1,0.6}
\definecolor{Skin}{rgb}{1,0.71,0.69}
\definecolor{Grey}{rgb}{0.5,0.5,0.5}
\definecolor{LightGrey}{rgb}{0.6,0.6,0.6}
\definecolor{Black}{rgb}{0,0,0}
\definecolor{White}{rgb}{1,1,1}
\newcommand{\red}{\color{Red}}
\newcommand{\green}{\color{Green}}
\newcommand{\blue}{\color{Blue}}

\newcommand{\grey}{\color{Grey}}

\newbool{blind}
\setbool{blind}{false}

\ifbool{blind}
{\AtBeginEnvironment{blind}{\comment}%
 \AtEndEnvironment{blind}{\endcomment}}{}

\DeclareMathOperator*{\argmin}{argmin}

\makeatletter
\DeclareRobustCommand\onedot{\futurelet\@let@token\@onedot}
\def\@onedot{\ifx\@let@token.\else.\null\fi\xspace}

\newcommand{\ie}{i.e.,\xspace}

\newcommand{\vs}{vs\onedot\xspace}
\makeatother

\newcommand{\xnearest}{x_{\text{nearest}}\xspace}
\newcommand{\xrand}{x_{\text{rand}}\xspace}
\newcommand{\xnew}{x_{\text{new}}\xspace}
\newcommand{\xinit}{x_{\text{init}}\xspace}
\newcommand{\xnext}{x_{\text{next}}\xspace}

\title{\LARGE \bf
Deep sequential models for sampling-based planning
}

\ifbool{blind}{}{%
  \author{Yen-Ling Kuo, Andrei Barbu, and Boris Katz%
    \thanks{This work was supported by the Center for Brains, Minds and
      Machines, NSF STC award 1231216, the Toyota Research
      Institute, CBMM-Siemens Graduate Fellowship, and the MIT-IBM Brain-Inspired Multimedia Comprehension
      project.}%
    \thanks{\protect\raggedright Computer Science and AI Laboratory, MIT
      {\tt\small \{ylkuo,abarbu,boris\}@mit.edu}}%
  }}

\usetikzlibrary{calc}
\usetikzlibrary{fit}
\usetikzlibrary{positioning}
\usetikzlibrary{decorations.markings}
\usetikzlibrary{shapes.geometric}
\usetikzlibrary{shapes.multipart}
\usetikzlibrary{bayesnet}
\usetikzlibrary{tikzmark}
\usetikzlibrary{backgrounds} 
\usetikzlibrary{shapes}
\usetikzlibrary{3d}
\usetikzlibrary{arrows.meta}
\usetikzlibrary{shapes.geometric}

\definecolor{pinegreen}{cmyk}{0.92,0,0.59,0.25}
\definecolor{royalblue}{cmyk}{1,0.50,0,0}
\tikzstyle{cblue}=[circle, draw, thin,fill=cyan!20, scale=0.8]
\tikzstyle{obs}=[circle, draw, thin,fill=gray!20, scale=0.8]
\tikzstyle{qgre}=[circle, draw, scale=0.8]
\tikzstyle{rpath}=[thick, black, opacity=0.4]

\tikzstyle{pathnode}=[circle, draw, scale=0.04]
\tikzstyle{graphnode}=[circle, draw, scale=0.15,ultra thin]
\tikzstyle{graphedge}=[-{Latex[black,length=0.15ex,width=0.15ex]}, shorten >= 0.04ex, ultra thin]
\tikzstyle{graphdashed}=[dash pattern=on 0.2pt off 0.1pt]

\newcommand{\bugtrap}%
{
  \draw[ultra thin, purple] (0,0,0)--(0,1,0) (0,1,0)--(1,1,0) (1,1,0)--(1,0,0) (1,0,0)--(0,0,0);
  \coordinate (a) at (0.25,0.25,0);
  \coordinate (b) at ($(a)+(0.21,0,0)$);
  \coordinate (c) at ($(b)+(0,0.24,0)$); 
  \coordinate (d) at ($(a)+(0,0.5,0)$); 
  \coordinate (e) at ($(d)+(0.5,0,0)$); 
  \coordinate (f) at ($(e)+(0,-0.5,0)$); 
  \coordinate (g) at ($(f)+(-0.21,0,0)$); 
  \coordinate (h) at ($(g)+(0,0.24,0)$);
  \draw[ultra thin, purple] (a)--(b) (b)--(c) (d)--(a) (d)--(e) (e)--(f) (f)--(g) (g)--(h);
}

\begin{document}

\maketitle
\thispagestyle{empty}
\pagestyle{empty}

\begin{abstract}
  We demonstrate how a sequence model and a sampling-based
  planner can influence each other to produce efficient plans and how such a
  model can automatically learn to take advantage of observations of the
  environment.
  Sampling-based planners such as RRT generally know nothing of their
  environments even if they have traversed similar spaces many times.
  A sequence model, such as an HMM or LSTM, guides the search for good
  paths.
  The resulting model, called DeRRT$^*$, observes the state of the planner and
  the local environment to bias the next move and next planner state.
  The neural-network-based models avoid manual feature engineering by
  co-training a convolutional network which processes map features and
  observations from sensors.
  We incorporate this sequence model in a manner that combines its likelihood
  with the existing bias for searching large unexplored Voronoi regions.
  This leads to more efficient trajectories with fewer rejected samples even in
  difficult domains such as when escaping bug traps.
  This model can also be used for dimensionality reduction in multi-agent
  environments with dynamic obstacles.
  Instead of planning in a high-dimensional space that includes the
  configurations of the other agents, we plan in a low-dimensional subspace
  relying on the sequence model to bias samples using the observed behavior of
  the other agents.
  The techniques presented here are general, include both graphical models and
  deep learning approaches, and can be adapted to a range of planners.
\end{abstract}

\section{INTRODUCTION}


When you navigate an environment containing new agents, obstacles, and goals, you
can rely on previous experiences to guide your actions.
Having seen similar agents before allows you to predict the motions of the ones
you encounter in the future.
Having seen obstacles, whether static or dynamic, allows you to efficiently
navigate around them.
Having seen certain goal types allows you to determine what preconditions must
be satisfied before meeting those goals.
In each case, your expectations about the future of the plan are conditioned on
your previous experiences, current plan, and local observations to help you
navigate.
This is the process we are modeling here.


\begin{figure}
  \centering
  \vspace{20ex}
  \begin{tikzpicture}[square/.style={regular polygon,regular polygon sides=4}]

    \begin{scope}[rotate around x=40,transform canvas={scale=5},yshift=-12,xshift=-10]
      \bugtrap
      \node[pathnode, square, scale=2, red, fill=red] (start) at ($(a)+(0.05,0.37,0)$) {};
      \node[pathnode, square, scale=2, Green, fill=Green] (end) at ($(a)+(-0.1,0.5,0)$) {};
      \node[pathnode, blue, fill=blue] (n1) at ($(start)+(0.06,-0.08,0)$) {};
      \node[pathnode, blue, fill=blue] (n2) at ($(start)+(0.14,0.02,0)$) {};
      \node[pathnode, blue, fill=blue] (n3) at ($(n1)+(0.024,-0.11,0)$) {};
      \node[pathnode, blue, fill=blue] (n4) at ($(n1)+(0.13,0.02,0)$) {};
      \node[pathnode, blue, fill=blue] (n5) at ($(n4)+(0.08,0.10,0)$) {};
      \node[pathnode, blue, fill=blue] (n6) at ($(n4)+(0.03,-0.05,0)$) {};
      \node[pathnode, red, fill=blue] (n7) at ($(n6)+(0.08,-0.03,0)$) {}; 
      \node[pathnode, Green, fill=blue] (n8) at ($(n6)+(-0.03,-0.07,0)$) {};
      \draw[ultra thin,blue] (start)--(n1) (start)--(n2) (n1)--(n3) (n1)--(n4) (n4)--(n5) (n4)--(n6);
      \draw[ultra thin, Green] (n6)--(n8);
      \begin{pgfonlayer}{background}
        \begin{scope}[transform canvas={scale=5}]
          \fill[opacity=0.5, fill=gray] (n6) circle (0.1);
        \end{scope}
      \end{pgfonlayer}
      \node[graphnode, above=1 of start, xshift=-10ex, yshift=0ex] (gstart) {};
      \node[graphnode, above=1 of n1, xshift=-1ex, yshift=0ex] (gn1) {};
      \node[graphnode, above=1 of n2, xshift= 1ex, yshift=5ex] (gn2) {};
      \node[graphnode, above=1 of n3, xshift=-1ex, yshift=-5ex] (gn3) {};
      \node[graphnode, above=1 of n4, xshift=2ex, yshift=0ex] (gn4) {};
      \node[graphnode, above=1 of n5, xshift=8ex, yshift=5ex] (gn5) {};
      \node[graphnode, above=1 of n6, xshift=10ex, yshift=-5ex] (gn6) {};
      \draw[graphedge, graphdashed] (gstart) -- (start);
      \draw[graphedge, graphdashed] (gn1) -- (n1);
      \draw[graphedge, graphdashed] (gn2) -- (n2);
      \draw[graphedge, graphdashed] (gn3) -- (n3);
      \draw[graphedge, graphdashed] (gn4) -- (n4);
      \draw[graphedge, graphdashed] (gn5) -- (n5);
      \draw[graphedge, graphdashed] (gn6) -- (n6);
      \draw[graphedge, graphdashed] (gn6) -- (n7);
      \draw[graphedge] (gstart) -- (gn1);
      \draw[graphedge] (gstart) -- (gn2);
      \draw[graphedge] (gn1) -- (gn3);
      \draw[graphedge] (gn1) -- (gn4);
      \draw[graphedge] (gn4) -- (gn5);
      \draw[graphedge] (gn4) -- (gn6);
    \end{scope}
  \end{tikzpicture}
  \vspace{13ex}
  \caption{A DeRRT$^*$-based planner starts {\red at the red square} and tries
    to reach the {\green green square} while escaping from a bug trap.
    The {\blue search tree, shown as blue circles}, is mirrored by a sequence
    model, an HMM or LSTM\@.
    When expanding the tree, a free-space sample is drawn, steered toward, and the {\red
      resulting node, shown as a red circle}, is used to find the closest
    node in the tree; as in RRT\@.
    The sequence model, with state corresponding to that closest node, observes
    this free-space sample, the path leading to this node,
    along with
    {\grey local visual or map features, shown in gray}, and predicts a
    {\green modified direction, shown in green}, which is then connected
    to the search tree.
    A new state for the sequence model is also predicted and connected.
    This process incorporates the bias to explore free space of RRT-based
    planners with a co-evolving sequence model and observations of the
    environment.}
  \label{fig:overview}
\end{figure}
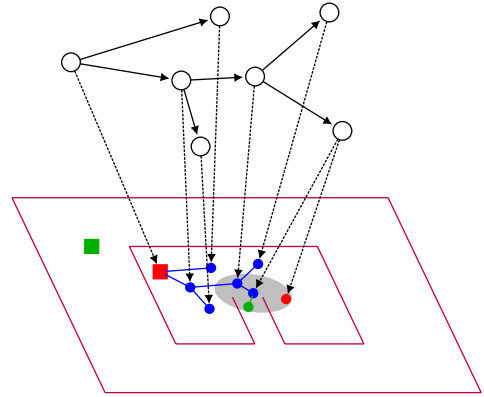

Existing sampling-based planners have difficulty taking advantage of such
information.
Most planners, like RRT$^*$~\cite{Karaman2011:rrtstar}, sample uniformly and take no
heed of the environment.
RRT$^*$, Rapidly-exploring Random Tree, is part of a family of algorithms
\cite{elbanhawi2014sampling,gammell2015batch,burget2016bi,adiyatov2017novel}
that explore a configuration space by sampling moves while avoiding invalid
states.
Dynamic environments, in particular, pose many challenges.
They combine uncertain sensing of the position of obstacles and agents with
uncertainty about the future path of those obstacles and the actions being
performed by other agents.
To improve planning in these domains, we adopt a set of techniques from computer
vision.
We bias the growth of the RRT$^*$ search
tree~\cite{Urmson2003:HeuristicBiasRRT,Lindemann2004:VoronoiBiasRRT} given prior
experience and a sensed environment.
Hidden Markov Models (HMMs)~\cite{baum1966statistical}, and stacked LSTMs
\cite{hochreiter1997long} are powerful activity recognizers
\cite{siddharth2014seeing,yu2015compositional,donahue2015long} but so far they have seen
little use in improving robotic planning.
We demonstrate how to adapt such sequence models to robotic planning using a
general approach that can employ either graphical models or neural networks.

A sequence model co-evolves alongside a sampling-based planner as shown in
Fig.~\ref{fig:overview}.
At each planning step, both the planner and the sequence model are stepped
forward while the next sample from the planner is conditioned on the sequence
model.
That sequence model can observe local features of the environment as well as the
current plan to provide good samples.
Moreover, we can avoid feature engineering and learn the
relevant features of the environment by co-training a convolutional network (CNN) with the LSTMs.
We refer to this algorithm as DeRRT$^*$, for deep RRT$^*$, although the
techniques presented here can be adapted to other sampling-based planners.

Several prior approaches have considered guiding planners with local
observations of static and dynamic obstacles.
\citet{Fulgenzi2008:GPRRT} demonstrate how a Gaussian process can be combined
with RRT to update plans conditioned on the observations of the motions of
dynamic agents.
This model estimates the positions of static and dynamic agents and assumes that
the velocity and direction of motion are constant.
Like the approach presented there, we plan incrementally and sample from the
stochastic model at every timestep.
Unlike this earlier work, we model the uncertainty in observing the position of
obstacles by relying on the ability of the HMMs and LSTMs
\cite{barbu2012simultaneous,ramanathan2016detecting} to capture this notion.
Additionally, we also allow for obstacles and agents that change velocity or
direction while including observations of features of those obstacles and agents.
Features are learned entirely automatically in the case of LSTMs.
Capturing the dynamics and appearance features of other entities can be helpful
in predicting future behavior and enhancing planner performance.
\citet{Aoude2013:DynamicObstacles} extend the work above to include a simulator
which further constrains the possible trajectories of other agents.
Such simulation is a natural extension to the models presented here using a
range of probabilistic programming approaches~\cite{Le2017:InferenceCompilation}
developed for computer vision~\cite{kulkarni2015picture} and
robotics~\cite{narayanaswamy2012seeing}.

The closest work to our approach is that of
\citet{Bowen2014:HMMTaskModel,bowen2016asymptotically}.
Their approach learns an HMM model for the trajectory of plans which achieves a
task and performs exhaustive inference in the cross-product space of that HMM
and the configuration space of the planner.
That work considers only domains where bidirectional planning is possible; we do
not rely on such information here.
Additionally, it only considers sequence models which observe the state of the
configuration space and one additional feature, the distance between landmarks
and the end effector.
This prevents modeling the time-varying motion of other agents, although by
detecting the arrival of new landmarks using a manually-set threshold, their
approach can replan when the environment changes.
This process does not account for perceptual uncertainty as to the position or
even presence of objects.
The inference algorithm considered in that work relies on a discrete HMM state.
The approach presented here can employ either continuous HMMs or arbitrary deep
learning sequence models such as stacked LSTMs or GRUs.

Similarly to the work above, \citet{Kim2018:PlanningGAN} use a generative
adversarial network, GAN~\cite{GAN}, to learn an action sampler.
At each timepoint, they sample a new action conditioned on the learned model.
Unlike our approach, the GAN does not model the dynamics of other
objects or capture uncertainty when sensing.
\citet{janson2018monte} show a Monte-Carlo planning approach that, like the
model described here, incorporates uncertainty in the observation of obstacles.
It does not, however, consider the dynamics of obstacles or learn to extract
relevant local features automatically.
\citet{arslan2017sensory} modify the steering function of a sampling-based
planner, in their case PRM, to include sensory information.
Unlike our approach, their work does not include a sequence model to
model the dynamics of obstacles or other agents, or automatically learn to
extract features from the environment.

Encoding the dynamics of other agents provides a major advantage: efficiently
reasoning in multi-agent scenarios.
Nominally the presence of other agents makes the reasoning problem exponentially
harder.
One must reason both about one's configuration space and about the configuration
space of the other agents.
In the case of a single robotic arm manipulating an object,
\citet{schmitt2017optimal} show how, without needing any training, one can reduce
the configuration space.
Our method would be useful in this domain as it is more general at the expense
of requiring training data.
Work by \citet{vcap2013multi} and \citet{Chen2015:MultiagentSeqConvex}
demonstrates how planners, including sampling-based planners, can be adapted to
multi-agent environments and cooperative planning without this exponential
slowdown.
Rather than cooperatively learning to plan, we demonstrate an agent avoidance
task which relies on the sequence model to learn to avoid the other agents.
\citet{kiesel2017effort} consider a related problem, learning to perform
kinodynamic planning.
Here we do not consider learning the dynamics of the robot being driven, instead
only focusing on other agents and objects, but this would be an interesting
direction for future work.





This work makes a number of contributions.
We show how to combine sequence models with sampling-based planners in a manner
that incorporates either graphical models or neural
networks.
Learned features of the trajectory, the local map, and the obstacles are combined
together to improve the generated plans in novel environments.
By borrowing its structure from computer vision algorithms for object tracking
and activity recognition, the model incorporates perceptual uncertainty.
The resulting model captures the dynamics of other agents to plan in multi-agent
scenarios.
Additionally, the model we present is flexible and can easily be adapted to new
sampling-based planners.
We demonstrate DeRRT$^*$ with a classical narrow passage, a bug trap scenario,
that co-trains a CNN to encode environmental cues, and a
multi-agent scenario where we take advantage of the learned patterns of motion
of the other agents.
We expect that by employing general-purpose sequence models, which have seen
great success in natural language processing and computer vision, to planning, this approach not only improves
performance but opens the door to further cross-pollination between these areas.


\section{PLANNING WITH SEQUENCE MODELS}

The algorithm presented here, DeRRT$^*$, combines a sequence model with a
sampling-based planner, RRT$^*$.
RRT-based algorithms create a tree which explores a configuration space.
The tree is used to efficiently connect an initial state to a goal state in that
configuration space.
Given an initial state, RRT$^*$ samples locations uniformly and then attempts to
connect them to that original node.
The tree reaches outward to cover the configuration space with a bias for large
unexplored Voronoi cells, eventually finding paths to the goal state.
In this way, RRT interleaves two steps: picking a point and steering toward it
while avoiding obstacles and infeasible areas.
See algorithm~\ref{alg:rrt} for a prototypical RRT\@.
RRT$^*$ is an asymptotically optimal version of RRT~\cite{Karaman2011:rrtstar}.
There are a number of common extensions to RRT, for example, bidirectional RRT
which considers the goal in addition to the initial state.
The work presented here can easily be adapted to such enhanced RRTs.

\begin{algorithm}
\caption{A prototypical RRT algorithm.}
\label{alg:rrt}
\begin{algorithmic}[1]
\State $V \gets \{\xinit\}; E \gets \emptyset$
\For{$1 \dots n$}
  \State $\xrand \gets \mathtt{SampleFree}()$
  \State $x_\text{nearest} \gets \mathtt{Nearest}(G=(V,E), \xrand)$
  \State $\xnew \gets \mathtt{Steer}(x_\text{nearest}, \xrand)$
  \If{$\mathtt{ObstacleFree}(\xnearest, \xnew)$}
    \State $V \gets V \cup \{\xnew\}$
    \State $E \gets E \cup \{(x_\text{nearest}, \xnew)\}$
  \EndIf
\EndFor
\State \Return $G = (V, E)$
\end{algorithmic}
\end{algorithm}

\subsection{RRT$^*$ with sequence models}

At each iteration of the RRT algorithm, we simultaneously extend the tree and a sequence
model.
Just as RRT trees have a branching structure, the sequence model will have that
same branching structure.
This conditions future states of the sequence model on past states for that
particular hypothesized plan.
Fig.~\ref{fig:overview} shows an example of this process.
As a node is expanded, the precise position in the configuration space of a new
candidate node is sampled from the sequence model.
To implement this, we modify the steering function to move in a direction given
by the sequence model while conditioning it on the current state, the desired
free space direction, and observations of the local environment around the
current state.

While several extensions to RRT consider changing the sampling function, here we
instead change the steering function.
This distinction is important and we take this approach for several reasons.
\begin{compactenum}
\item It preserves the most desirable property of RRT, its bias for large
  Voronoi regions.
  Exploring novel regions helps in difficult domains where simply attempting to
  directly reach the goal is unlikely to succeed.
  At the same time, if one wants to incorporate the goal position, this approach is
  easily extended to bidirectional RRT where two trees grow toward each other: one tree from the initial state and one tree from the goal state.
\item There is no need to change the sampling function.
  If the sequence model for the steering function has high confidence, the random sampled
  direction in free space is irrelevant.
  %
  %
  Intuitively, the sequence model controls how much exploration \vs exploitation
  is occurring based on its confidence in the next direction.
\item We would like to take advantage of local observations to help guide the
  algorithm.
  To do this, we allow the sequence model to observe those features, and in the
  case of the deep learning approach, to learn the nature of those
  features.
  Steering moves are small, making local decisions about the direction of
  motion, while free space-sampling controls the overall direction of motion.
  Local features are far more informative for small moves than for deciding what
  the overall direction across an entire map or maze might be.
\end{compactenum}

We keep the overall structure of RRT$^*$\citep{Karaman2011:rrtstar} unchanged 
modifying two of its component functions, $\mathtt{Steer}$ and
$\mathtt{Rewire}$.
As described above, we modify the $\mathtt{Steer}$ to guide the planner toward a
direction informed by the sequence model instead of just minimizing the distance
to the uniformly sampled node $\xrand$.
For performance reasons, this necessitates an update to $\mathtt{Rewire}$ to
cache the state of the sequence model when a node changes its parent.

When steering, one starts from node $x_\text{nearest}$ and heads in the direction
of the sampled point $\xrand$.
The end node of a single step of the steering function, $\xnew$, lies within a
distance $r$ of $\xnearest$, within a sphere $\mathcal{B}_{\xnearest, r}$.
In the original RRT$^*$, $\xnew$ is chosen to minimize the distance to
$\xrand$.
We replace this function with $\mathtt{SteerWithModel}$, as shown in
algorithm~\ref{alg:steer}.

$\mathtt{SteerWithModel}$ proceeds as follows.
First, we find $\mu$, the optimal point according to the original RRT$^*$
algorithm.
Next, we would like to sample a point within steering distance $r$ of $\xnearest$
conditioned on the sequence model, $\lambda$, along with any observations from
available sensors, $\mathtt{Obs}$.
When the sequence model allows for efficient conditioning of the samples based
on this sphere and sensor data, we can directly sample from the posterior.
Practically, most models do not allow for this and we instead sample a fixed
number of points, $k$ points, compute the likelihood of each, and sample
proportionally to those likelihoods.
Other approaches to drawing samples for the steering function such as
Monte-Carlo methods would also be appropriate but we intend to draw few samples
in a small region meaning that the advantages of such approaches are outweighed
by their additional runtime.



\begin{algorithm}
\caption{$\mathtt{SteerWithModel}(\xnearest,\xrand)$}
\label{alg:steer}
\begin{algorithmic}[1]
\State $\mu \gets \displaystyle\argmin_{z \in \mathcal{B}_{\xnearest, r}}\lVert z - \xrand \rVert$
\State $P \gets \emptyset$
\For{$1\ldots k$}
  \State $\xnext \gets \mathtt{SampleUniform}(\xnearest, \mu, r)$
  \State $p_\text{next} \gets P(\xnext, \mu, \mathtt{Obs}|\lambda, \xinit, \dots, \xnearest)$
  \State $S \gets S \cup \{(\xnext, p_\text{next})\}$
\EndFor
\State ($\xnew, p_\text{new}) \gets \mathtt{Sample}(S)$
\State \Return $\xnew$
\end{algorithmic}
\end{algorithm}


Intuitively, when the sequence doesn't provide any information about the
configuration space, it can learn to simply provide high likelihood when the hypothesized direction $\xnext$ is close to $\mu$.
This reverts the steering function to the one from the original RRT$^*$.
At the other extreme, the sequence model may choose to disregard the free space
samples if the future path is clear.
The structure of most sequence models makes computing this likelihood term very
efficient by decoupling the cost into the cost of the previous path, which is
shared by all future paths, and an additional term for the new position and the
new observations.
Next, we describe how this generic model is instantiated in the case of HMMs and
then neural networks.

\subsection{Steering with HMMs}

HMMs are easy to train, they do not generally require much data, and usually provide
efficient exact inference algorithms.
They do on the other hand require feature engineering.
Regardless of the domain we always include a feature modeled by a normal
distribution which observes the difference between the hypothesized direction,
$\xnext$, and the optimal direction according to the original RRT$^*$, $\mu$.
This allows the HMMs to elect to follow the original steering function.
Nominally, it is possible to also recover this original behavior when the HMMs
assign equal likelihood to all outcomes but this can be difficult to learn.
We extract other features from the environment and current trajectory and also
model them using normal distributions.

We only consider HMMs with a finite number of discrete states although the
inference algorithm only requires that a likelihood of a path is computable.
It is agnostic to the number or type of states since we marginalize over all
states.
To do this efficiently, we employ the forward algorithm.
We take advantage of the Markov property to decompose the likelihood function
into one computed for the existing path, $P(\xinit \dots \xnearest|\lambda)$, and
an additional update term, $P(\xnearest,\xnext,\mu,\mathtt{Obs}|\lambda)$.
The Markov property also allows us to efficiently cache only the final entry in
the lattice at each node of the RRT tree.
Training the model employs an EM algorithm along with a collection of traces.

HMMs have more important limitations than just requiring feature engineering.
They have difficulties capturing complex temporal dynamics because of an implied
exponential state duration model; efficient algorithms are scarce for other
state duration models.
Additionally, the Markov property also limits the complexity of the dynamics that
can be modeled without an explosion in the number of states.

\subsection{Steering with neural networks}

These limitations of HMMs prompt us to employ neural networks instead.
Using recurrent networks to approximate complex probability distributions is not
new.
For example, \citet{Le2017:InferenceCompilation} show that probabilistic
programs can be compiled into neural networks that take observations as input
and learn to perform inference.
One could accurately approximate the HMMs with the neural networks using such
techniques.

We consider classes of recurrent models such as RNNs~\cite{hopfield1982neural},
LSTMs~\cite{hochreiter1997long}, and GRUs~\cite{cho2014properties}.
While we often for shorthand reasons refer to LSTMs, the planning algorithm is
intentionally agnostic to the specifics of the chosen recurrent model.
In each case, these algorithms have a state that is propagated at each time step,
Unlike HMMs, states are not necessarily discrete or interpretable.
At each time step, models observe the same quantities the HMMs do.
And like with the HMMs, we use the outputs of the recurrent models to compute a
likelihood for each direction in the steering function.

The recurrent networks take a local observation around the current
point, the current position, and the current optimal
direction.
Local observations and map features are embedded into a fixed-dimensional input
vector.
Convolutional layers can be co-trained with the recurrent model to
take input images of the map or any other perceptual information.
This eliminates the need for feature engineering and provides robustness to
perceptual uncertainty; we do not need to commit to the presence or absence of a
feature in the environment.
The network learns one embedding layer to embed the local observations at
the current position.
In addition, at each time step, the previous state \textemdash\ an arbitrary
vector \textemdash\ is propagated and both a new state and a new direction are
produced.
The network is initialized with a zeroed state vector.

The recurrent models can in principle directly score a future state.
In practice, we found that having an explicit mixture model that combines the
optimal direction per the original RRT$^*$ with the direction preferred by the
LSTM results in models which are easier to train.
It also provides a level of interpretability to the model.
At each time step, we use the recurrent network, a step of which is evaluated by
the function $\eta$, to produce a mean and covariance matrix for a normal
distribution from which new directions can be sampled.
The likelihood of a steering move is computed as a mixture of
\begin{equation}
      \label{eq:nn-evaluation}
      q(x_t|\eta(x_{t-1}, \mathbf{s}_{t-1}, \mathtt{Obs}, \phi))
\end{equation}
and the likelihood of following the RRT$^*$ direction, $\mu$, is computed as a
normal distribution $N(\mu, \sigma)$ where $q$ is normal proposal distribution,
$\eta$ is the recurrent model (a function returning a mean direction and a
covariance matrix), $\mathbf{s}$ is the state vector of the model,
$\mathtt{Obs}$ is an embedding of the observation vector, and $\phi$ are the
parameters of the recurrent model.
In practice the value of $\sigma$ is not important, the network learns to
compensate for the chosen value, but its addition makes for a coherent
stochastic model.
For efficiency, similarly to the HMM case, we store the current state of the
recurrent network at each node in the search tree and incrementally compute the
likelihood of a path.

At training time, the network is supplied with a series of traces of successful
plans.
While we do not do so here, we could include unsuccessful plans as part of the
training set for the neural networks and even the HMMs by using discriminative
training.
At each time step during training, stochastic gradient descent is used to maximize
the likelihood shown in equation~\ref{eq:nn-evaluation}.
In essence, we have samples from an $n$-dimensional normal distribution, where
$n$ is the size of the configuration space, along with the network which produced
the mean and covariance matrix from which these samples were drawn
at each time step.
We then train this network to maximize the likelihood of the observed sequences.

\subsection{Multi-agent planning}\label{sec:multi-agent-methods}

Sequence models can capture the dynamics of other agents.
Doing so is useful for both avoiding moving obstacles, such as pedestrians, and
for coordinating with those agents, such as merging into traffic with other
autonomous cars.
Coordination problems such as these are often computationally difficult for each
individual agent, particularly when perception is unreliable.
On the other hand, simulating the correct behavior of multiple agents when a
shared oracle is available is easy.
This is a good fit for the type of planning performed by DeRRT$^*$.
Recurrent models have sufficiently large capacity to recognize
the actions of other agents and learn a multi-agent plan directly.
Simulations provide large amounts of training data to tune these recurrent
models.
Intuitively, rather than attempting to plan in a large configuration space that
is a cross-product of motions of both the robot and the other agents, we plan in
a subspace, the robot's own motions, and rely on the sequence models to perform
a type of dimensionality reduction.
The space of the robot's motions is warped to account for the motions of the
other agents.

We augment the model to explicitly include observations of other agents and to
explicitly reason about them.
At each time step, an embedding of the sensor information for each agent is
produced and also provided to the network.
Nominally, one could encode this information by taking as input a sequence of
observations, one for each interacting agent, embedding each individually and
using a separate sequence model embedding all information for all agents into a
single vector.
This is similar to how sentences are embedded into vectors using embeddings for
each word and a recurrent model to combine that per word
embeddings~\cite{palangi2016deep}.

However, such models can be difficult to train.
Here, we consider a complementary approach that takes advantage of the
probabilistic interpretation of the sequence models described above.
For each agent observed, we predict a mean and covariance matrix for the
direction to steer in.
Agents that do not influence the current plan will have uninformative directions
while critical agents will have highly concentrated distributions.
One could also compute directions for pairs or triples of agents in this manner
if the relationships between agents, not just between the robot and a single
agent, are important for coordination.
This process is also data efficient: a single example of multiple cooperating
agents leads to $2 \times \textit{number of agent pairs}$ training
examples as network weights are shared between all agent pairs.
At inference time, the steering direction is a mixture of the optimal RRT$^*$
direction, the direction according to the local observations of the model, and
the directions predicted for each observation of each agent.
This process efficiently scales to a large number of agents.

\section{EXPERIMENT RESULTS}

We tested DeRRT$^*$ in a number of challenging environments.
The sequence models described above were implemented in PyTorch and integrated
with the Open Motion Planning Library, OMPL~\cite{Sucan2012:OMPL}, using the
provided Python bindings.\footnote{Source code is available at https://github.com/ylkuo/derrt.}
The selected planning environments demonstrate three key features of the
approach presented here: planning efficiently in difficult domains such as
narrow channels, using local perceptual features to learn to escape a bug trap,
and multi-agent navigation that relies on learning the other agent's motion
patterns for coordination.

\subsection{Long narrow passage}\label{sec:narrow-passage}

Narrow passages pose a problem for sampling-based
planners~\cite{Hsu1997:NarrowPassage}.
The volume of a narrow passage is much smaller than the volume of the free space
making it hard to find good directions to move in.
Prior work has shown that the convergence of RRT-like algorithms depends on the
thickness of the narrow passage.

\begin{figure}[t]
\centering
\subfigure[Map with a DeRRT$^*$ tree and solution]{\includegraphics[width=0.28\textwidth]{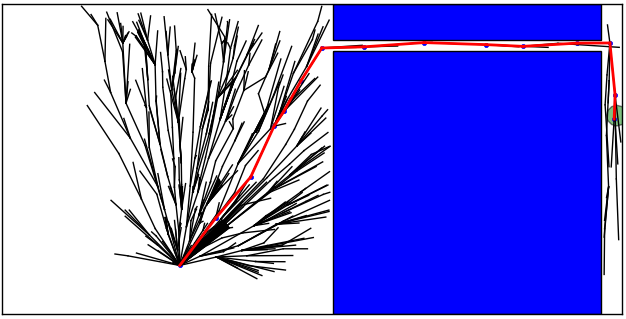}\label{fig:narrow_passage_example}}\\
\subfigure[DeRRT$^*$]{\includegraphics[width=0.22\textwidth]{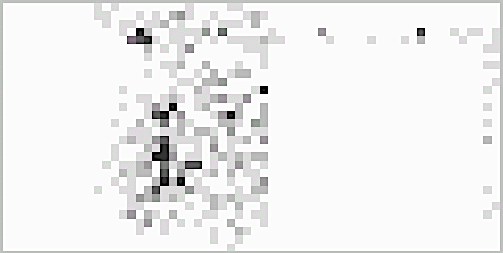}\label{fig:narrow_passage_heatmap_hmm}}
\subfigure[RRT$^*$]{\includegraphics[width=0.22\textwidth]{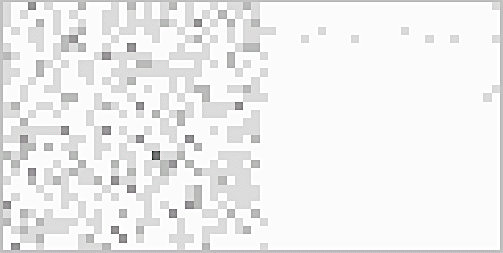}\label{fig:narrow_passage_heatmap_uniform}}
\caption{A test case for the long narrow passage.
  (a) The DeRRT$^*$ search tree, black, and solution, red.
  (b) A heat map of the DeRRT$^*$/HMM search tree.
  Note the distribution bias toward the channel entrance in the free space, efficient traversal of the channel,
  and more samples after the channel exit.
  (c) A heat map of the RRT$^{*}$ search tree.
  Note the high proportion of time spent looking for the entrance,
  less efficient traversal of the channel,
  and few samples in the second free space looking for the goal.
  Pure black represents allocating 1\% of the samples inside the given cell with linear
  interpolation to pure white.}
\label{fig:area_explored_passage}
\end{figure}

We created a 2D environment where a robot navigates from a start location to a
goal region.
At each instance of this problem, we randomized the start and end positions.
We uniformly sample the thickness and the length of the passage and place its
opening at a randomly sampled position.
Fig.~\ref{fig:narrow_passage_example} shows an example map.
The model was trained with 200 example sequences.

For the case of the HMMs, we had 3 hidden states, perhaps interpretable as corresponding to
having not yet entered the the passage, being in the passage, and having traversed it.
We extracted the agent's distance to the passage entrance, the presence of the
agent in the passage, and the coordinates of the agent on the map as observed
features.
Each feature was assumed to be independent from the rest by the use of a
diagonal covariance matrix.
The neural network sequence models used a GRU with an input layer to embed
these local features, two hidden layers with 32 dimensions, and a linear
proposal layer.

While at training time the map was always 300 by 300, at test time the map was
always 600 by 300.
%
%
We also significantly narrowed the passage size at test time.
This both makes the problem more challenging and makes the test examples
different from the training examples showing the generalization capabilities of
the sequence models.
Fig. \ref{fig:narrow_passage_example} shows a test case along with expanded
search tree and the final path.
%
%
There is no configuration similar to this one in the training examples.

\begin{table}
  \centering
  \caption{Success rates for the long narrow passage problem.}%
  \vspace{-2ex}
    \begin{tabular}{cccc}
      & RRT$^*$ & DeRRT$^*$/HMM & DeRRT$^*$/GRU \\
      success \% & 3.83\% & 24.02\% & 47.67\% \\
      standard deviation   & 1.68  & 3.74  & 4.38 \\
    \end{tabular}
    \vspace{-1ex}
    \begin{flushleft}
    A test set with different statistics than the training set was used drawing
    600 samples.
    Note the much higher success rate of DeRRT$^*$ with either HMMs or GRUs.
    While the standard deviation is also somewhat higher, it is miniscule
    compared to the difference in performance between the approaches.
    \end{flushleft}
  \label{tbl:success_rate_passage}
\end{table}

At test time, we ran each of RRT$^*$, DeRRT$^*$ with HMMs, and DeRRT$^*$ with GRUs
for 500 rounds with 600 sampled nodes in each round.
DeRRT$^*$ was able to succeed far more reliably with such small number of
samples,
$24\%$ to $47\%$ of the time, a 6 to 12-fold increase, depending on whether the
HMMs or the GRUs-based models were used.
Table~\ref{tbl:success_rate_passage} summarizes the success rate of each
planner.
We emphasize that the test set is disjoint from the training set both in terms
of individual examples but also in terms of the statistics of the examples; the
narrower passages required the planners to generalize.

We would like for the algorithms to not just have high performance but also to 
behave in a directed manner, while still exploring free space.
Fig.~\ref{fig:narrow_passage_heatmap_hmm}
and~\ref{fig:narrow_passage_heatmap_uniform} summarize the heat map of the
search trees on the given test example.
Intensity corresponds to the number of times the tree visited that
region.
One can see that DeRRT$^*$ trades off exploration vs exploitation.
It still searches the free space but does so toward the channel entrance, more easily
traverses the channel, and samples more densely in the free space after the channel 
where the goal is.

\subsection{Bug trap}\label{sec:bug-trap}

\begin{figure}[t]
\centering
\subfigure[Bug trap]{\includegraphics[width=0.14\textwidth]{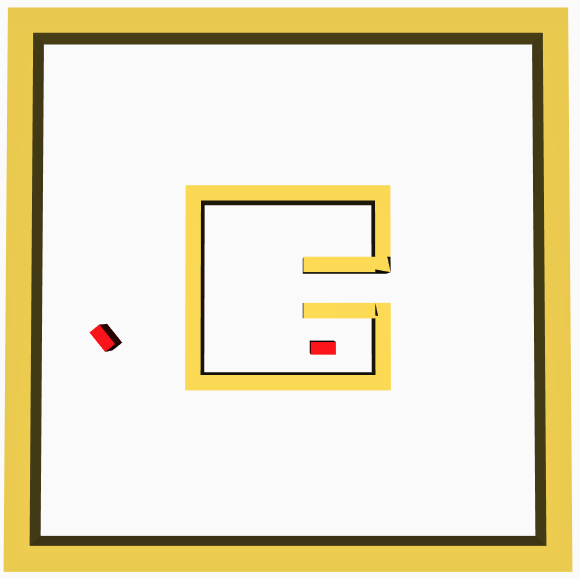}\label{fig:bugtrap_map}} \\
\subfigure[DeRRT$^*$]{\includegraphics[width=0.16\textwidth]{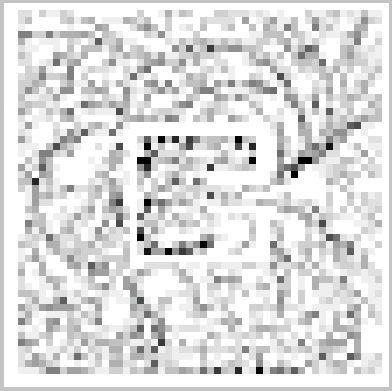}\label{fig:drrt_area}}
\subfigure[RRT$^*$]{\ \ \includegraphics[width=0.16\textwidth]{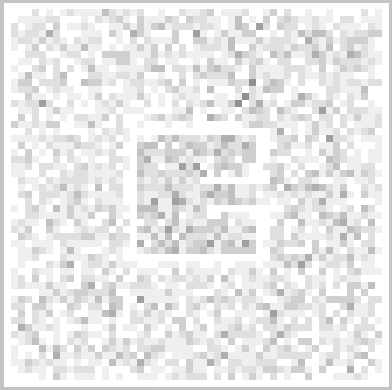}\label{fig:rrt_area}}
\caption{(a) The default OMPL bug trap example with its start and end position.
  Note the difficult central passage and dead-ends on either side of it.
  (b) A heat map of the DeRRT$^{*}$/GRU search tree.
  It efficiently learns to exit the trap and quickly
  focuses on sweeping in large arcs to locate the goal.
  (c) A heat map of the RRT$^{*}$ search tree.
  It spends more time in the trap and less time on finding the goal in the free
  space.
  Pure black represents allocating 0.2\% of the samples inside the given cell with linear
  interpolation to pure white.}
\label{fig:area_explored_bugtrap}
\end{figure}

The bug trap is a standard benchmark in OMPL.
It requires that a 2D robot escape from an inner chamber through a narrow
passage and then reach a goal in a large free space; see
Fig.~\ref{fig:bugtrap_map}.
This is made particularly hard by the shape of the exit which includes two dead ends.
Most samples in the free space will lead to steering into these areas.
For planners to escape from the bug trap, particularly sampling-based planners,
a very deliberate sequence of samples must be drawn to guide the robot to the
passage and then down the passage without getting trapped.

Unlike the narrow passage problem, the bug trap has much more distinctive local
visual features, \ie the shape of the trap.
To quickly escape, one has to recognize not just the presence of a gap but the
particular features of the central channel.
In this experiment, we cotrain the convolutional layers of the neural-network
sequence models to learn to recognize the presence of relevant map features
in order to reach a goal.
This eliminates the need for feature engineering or any other annotation aside
from a series of prior plans.

To again ensure that the training and test data are disjoint we manipulated the
example provided in OMPL\@.
%
%
We randomly rotated and translated the trap and randomly sampled the starting
position inside the trap and the goal configuration outside the trap.
Training samples were provided by running RRT$^*$ for 10000 planning steps on each problem
instance.
In total, we collected 1000 training sequences.

Instead of manually engineering features, we take as an observation
$21 \times 21$ local patch centered at the current node from a
$110 \times 110$-sized map.
The sequence model used was similar to that from
section~\ref{sec:narrow-passage}, with two modifications.
First, manually-engineered features were replaced with a two-layer convolutional
network, each layer containing a convolution followed by max polling.
The convolutions used $3 \times 3$ filters with 32 and 64 output channels respectively.
Max pooling used a $2 \times 2$ window with step size 2.
Second, we used GRUs instead of LSTMs as they proved easier to co-train
with the convolutional layers.

At test time, we compared RRT$^*$ to DeRRT$^*$.
Fig.~\ref{fig:bugtrap_map} shows the default bug trap from OMPL.
We ensure that the test and training sets are disjoint and that no motion
sequence in the training set solves a map from the test set.
Each planner is run with a maximum of 10000 planning steps.
%
%

\begin{figure}
\centering
\includegraphics[width=1.9in]{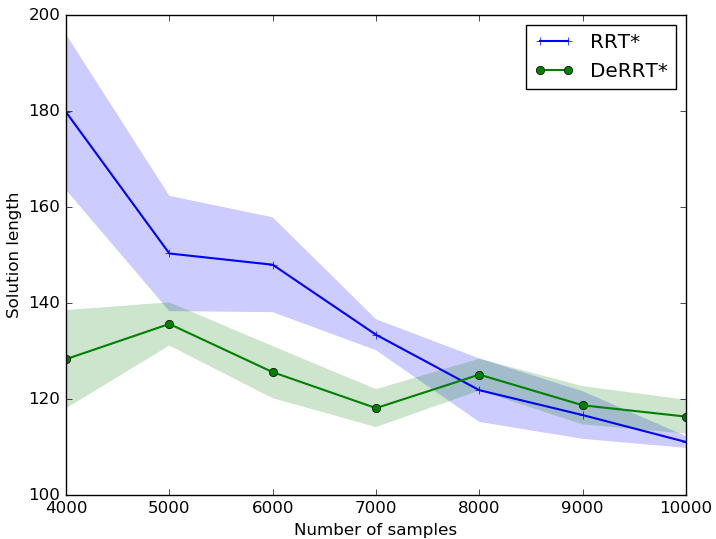}
\caption{Solution length as a function of the number of samples. DeRRT$^*$ is
  both more efficient and more stable.}
\label{fig:bugtrap_solution_length}
\end{figure}

\begin{figure}
\centering
\includegraphics[width=1.9in]{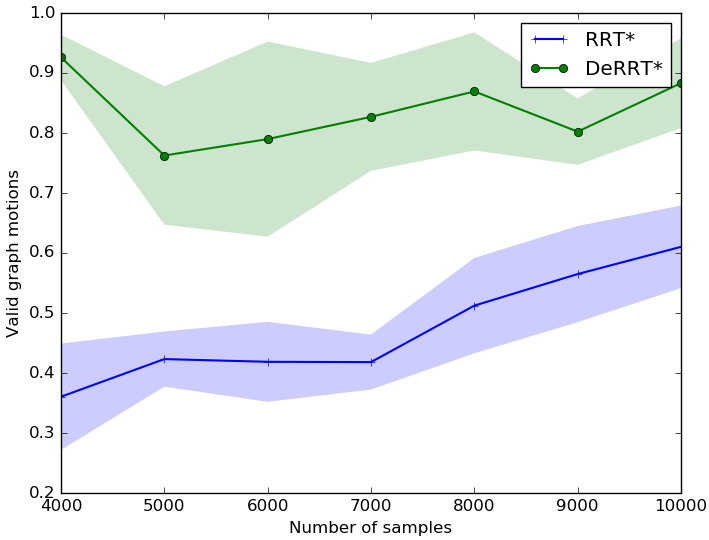}
\caption{Proportion of valid graph moves as a function of the number of
  samples. DeRRT$^*$ learns to avoid proposing invalid moves.}
\label{fig:bugtrap_graph_motions}
\end{figure}

Fig.~\ref{fig:bugtrap_solution_length} shows the solution length as a function
of the number of samples drawn.
Already by 4000 samples the sequence-model guided planner is performing as well
as RRT$^*$ with twice the number of samples.
The colored regions show 95\% confidence intervals.
Additionally, DeRRT$^*$ is more stable than RRT$^*$,
likely because the proportion of valid moves is around 0.8, as shown in
Fig.~\ref{fig:bugtrap_graph_motions}.
This makes DeRRT$^*$ far more efficient in its proposals when compared to RRT$^*$.
When considering more complex scenarios, such as an articulated robot with a complex mesh,
this can have an even more significant impact as the more expensive collision checking can
become a dominant concern in the runtime of sampling-based planners.

Figs.~\ref{fig:drrt_area} and ~\ref{fig:rrt_area} show heat maps computed the same
way as in the previous section where intensity is proportional to the density of nodes.
DeRRT$^*$ learns to sample less in dead ends and focuses on leaving the channel and exploring the outside space which contains the goal while RRT$^*$ tends to spend
much more time in the trap.

\subsection{Multi-agent coordination around a static obstacle}\label{sec:multi-agent}

\begin{table*}[ht!]
\caption{Comparing the success rate and path length as a function of the number
  of agents of four different planning approaches.}
\label{tab:ma_result}
\centering\begin{tabular}{ccccccccc}
 & \multicolumn{2}{c}{RRT$^*$} & \multicolumn{2}{c}{DeRRT$^*$/HMM} & \multicolumn{2}{c}{DeRRT$^*$/GRU} & \multicolumn{2}{c}{RRT$^*$-joint} \\ \cmidrule(lr){2-3} \cmidrule(lr){4-5} \cmidrule(lr){6-7} \cmidrule(lr){8-9}
  \# of agents & success & path length & success & path length & success & path length & success & path length \\[0.5ex]
2  & 0.82 & $116.74 \pm 2.82$ & 1.00 & $112.17 \pm 1.42$ & 1.00 & $\phantom{0}96.75 \pm 1.01$ & 0.98 & $169.51 \pm \phantom{0}9.11$ \\
4  & 0.68 & $122.84 \pm 3.46$ & 0.54 & $115.29 \pm 2.37$ & 0.90 & $110.88 \pm 1.81$ & 0.46 & $302.93 \pm 47.82$ \\
6  & 0.22 & $134.30 \pm 8.34$ & 0.30 & $111.74 \pm 2.22$ & 0.33 & $111.31 \pm 1.98$ & 0.40 & $300.79 \pm 89.39$ \\
8  & 0.26 & $140.45 \pm 6.46$ & 0.46 & $115.27 \pm 2.36$ & 0.18 & $108.37 \pm 2.05$ & 0.00 & *** \\
\end{tabular}
\end{table*}

\begin{figure}[t!]
\centering
\subfigure[]{\includegraphics[width=0.19\textwidth]{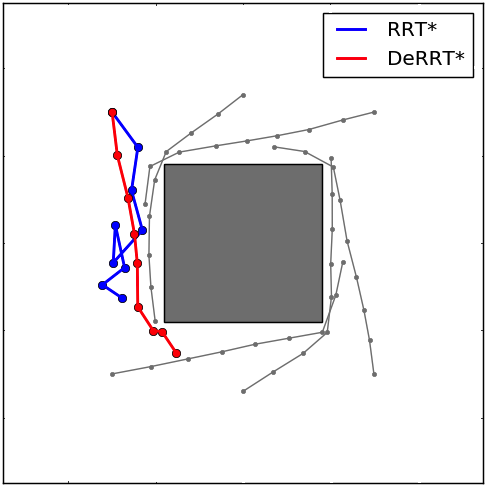}\label{fig:multiagent_half}}
\subfigure[]{\includegraphics[width=0.19\textwidth]{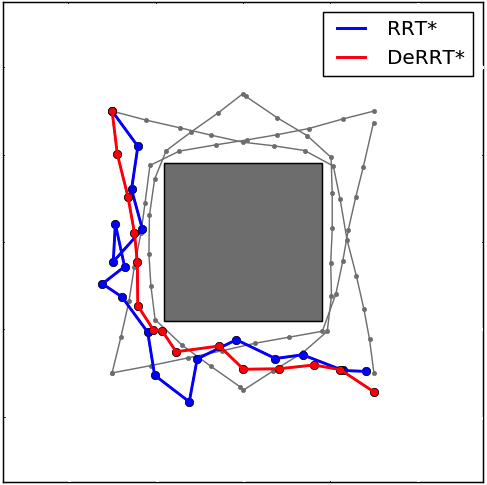}\label{fig:multiagent_full}}
\caption{Trajectory comparison in 6 agents swap position case at 500 samples per re-plan step. The grey lines are trajectories of other preloaded agents. (a) Trajectory at the mid-time (b) Trajectory at the end.}
\label{fig:trajectory_multiagent}
\end{figure}

Finally, we test the ability of the sequence models to learn to coordinate with
other agents using a task where agents must swap their positions around a
central obstacle while avoiding each other; see Fig.~\ref{fig:trajectory_multiagent}.
The obstacle makes the problem significantly harder than a standard position swap
in free space since moving randomly in free space makes it very unlikely
that agents will collide with each other.
To further increase the difficulty, we constrain all other agents to move
counter-clockwise around the obstacle.
This scenario is closely related to driving; one might view it as a
roundabout.
To scale the difficulty of the problem, we change the number of agents that must
avoid collisions.

Training data was generated by randomly placing an obstacle in a
$100 \times 100$ map, and up to four agents at random orientation around the
obstacle.
Motion sequences were supplied by running RRT$^*$ for 10,000 iterations.
This resulted in 200 different maps and 600 sequences in total.
We trained a 3-state HMM including the orientation and distances to the goal and
other agents.
Similarly to the previous sections, we trained a GRU-based planner including those
same features.
As described in section~\ref{sec:multi-agent-methods}, the neural-network-based
planner evaluates each of the different agents separately, proposes means and
covariance matrices for each, and draws a sample from the resulting mixture
model.

We compared DeRRT$^*$ to two other approaches, RRT$^*$ and RRT$^*$-join, with up
to eight agents.
RRT$^*$ only considers the configuration space of the robot while replanning at
each step.
It treats the map as providing snapshots of fixed obstacles, \ie the center 
block and the other agents, some of which happen to move between the snapshots.
RRT$^*$-joint plans in the joint configuration space of all agents.
By reasoning explicitly about the configuration space of other agents, it
can in principle take into account the expected trajectories of those agents.
Without a model of the behavior of the agents, RRT$^*$-joint has
difficulties making meaningful inferences aside from constraining
the likely paths of other agents to avoid imminent collisions, while at the same time 
incurring the cost of an exponentially increasing configuration space.

\begin{figure}
\centering
\includegraphics[width=1.9in]{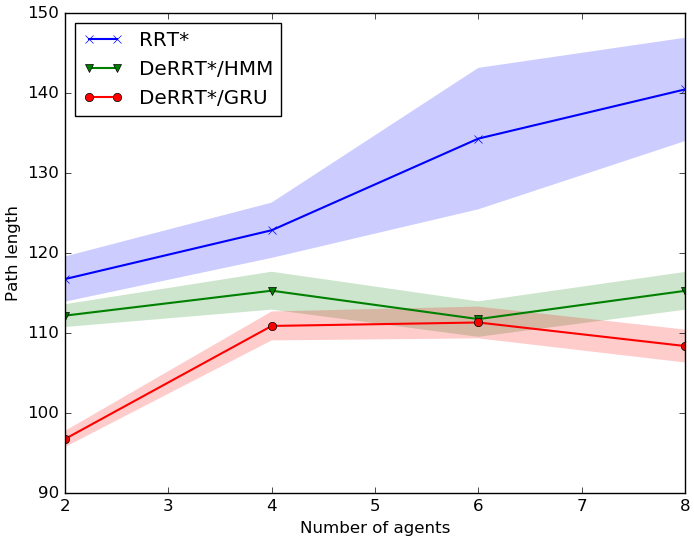}
\caption{Solution path length in multiagent navigation task}
\label{fig:ma_path_length}
\end{figure}

Table~\ref{tab:ma_result} shows detailed results comparing these four approaches
for a fixed number of samples per re-plan step, 100 for all except for RRT$^*$-join which due
to its exponentially larger search space requires 5000.
RRT$^*$-joint quickly degrades because even though it is nominally representing
the problem with higher fidelity than RRT$^*$, sampling in high dimensional spaces
is known to be very difficult.
DeRRT$^*$ with GRUs performs considerably better in terms of its success rate.
Note that DeRRT$^*$ was never trained on the same scenarios it was tested on and
it was never trained with 6 or 8 agents on the map, yet it seems to generalize well
to these new instances.
DeRRT$^*$ with either sequence model finds much shorter paths.
Fig.~\ref{fig:ma_path_length} shows the path length as a function of the
number of agents.
RRT$^*$-joint is omitted because its poor performance would make it difficult to
see the difference between any of the other algorithms.
As the number of agents increases, DeRRT$^*$ does not produce poorer paths,
although its likelihood of success does go down as finding collision-free paths
becomes more difficult.
Figs.~\ref{fig:multiagent_half} and~\ref{fig:multiagent_full} show an instance
of this problem along with full tracks, in grey, and partial tracks halfway through the execution, in color, for RRT$^*$ and DeRRT$^*$ solutions.
Even when both approaches reach the goal, the collision-free paths produced by DeRRT$^*$ are
much smoother.

\section{CONCLUSIONS}

We have introduced DeRRT$^*$, a sampling-based planner extending RRT$^*$ with
either a graphical model or a neural network in order to learn to plan more
efficiently.
The algorithm presented here is designed to accommodate a range of sequence
models, making minimal assumptions about either the graphical model or the deep
network.
This opens the door to using models that are successful in other areas, for
example, compositional models in vision or sequence-to-sequence models in natural
language.
We demonstrated how map features can be automatically learned with a
CNN and how, when relevant to a planning task, sequence models can learn to predict
the motions of other agents leading to more efficient plans.
In the future, we expect that planners which understand more about their environments,
perhaps by incorporating existing CNNs trained on large vision corpora, will navigate
more efficiently.
Similarly, sequence models which can capture existing knowledge about an environment
and reason about the consequences of actions may be better suited to carrying out
complex tasks.






\bibliographystyle{IEEEtranN}
\renewcommand*{\bibfont}{\footnotesize}
\bibliography{reference}

\end{document}